# Evaluating the fully automatic multi-language translation of the Swiss avalanche bulletin

Kurt Winkler[1], Tobias Kuhn[2], Martin Volk[3]

1 WSL Institute for Snow and Avalanche Research SLF, 7260 Davos Dorf, Switzerland.
2 Department of Humanities, Social and Political Sciences, ETH Zurich, Switzerland
3 Universität Zürich, Institut für Computerlinguistik, Binzmühlestrasse 14, CH-8050 Zürich

**Abstract.** The Swiss avalanche bulletin is produced twice a day in four languages. Due to the lack of time available for manual translation, a fully automated translation system is employed, based on a catalogue of predefined phrases and predetermined rules of how these phrases can be combined to produce sentences. The system is able to automatically translate such sentences from German into the target languages French, Italian and English without subsequent proofreading or correction. Our catalogue of phrases is limited to a small sublanguage. The reduction of daily translation costs is expected to offset the initial development costs within a few years. After being operational for two winter seasons, we assess here the quality of the produced texts based on an evaluation where participants rate real danger descriptions from both origins, the catalogue of phrases versus the manually written and translated texts. With a mean recognition rate of 55%, users can hardly distinguish between the two types of texts, and give similar ratings with respect to their language quality. Overall, the output from the catalogue system can be considered virtually equivalent to a text written by avalanche forecasters and then manually translated by professional translators. Furthermore, forecasters declared that all relevant situations were captured by the system with sufficient accuracy and within the limited time available.

**KEYWORDS:** machine translation, catalogue of phrases, controlled natural language, text quality, avalanche warning

## 1 Introduction

Apart from the requirements of being accurate and easy to understand, avalanche bulletins are highly time-critical. The delivery of up-to-date information is particularly challenging in the morning, when there is little time between incoming field observations and the deadline for publishing the bulletin – not enough time for manual translations or manual post-editing. For that reason, the new Swiss avalanche bulletin (Fig. 1) is generated by a fully automated translation system, which we have described in a previous publication [17]. Here we present evaluation results of this system after two winter seasons of operational use.

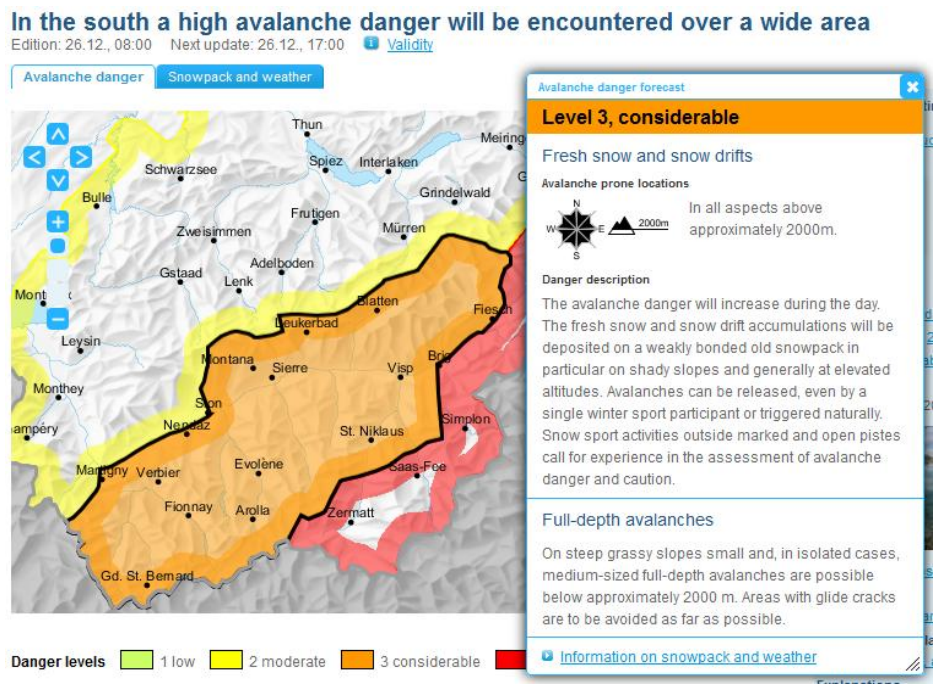

**Fig. 1.** Swiss avalanche bulletin. The danger descriptions originate from the catalogue of phrases (here in English). More examples and other languages are available at www.slf.ch (in summertime only as .pdf in the archive: www.slf.ch/schneeinfo/Archiv).

Despite the large effort on machine translation approaches and despite their promising results, the quality of fully automatic translations is still poor when compared to manual translations. For the publication of life-critical warnings when there is no time for proofreading or manual corrections, the reliability of existing translation systems is clearly insufficient.

For many years, the daily Swiss national avalanche bulletin was manually translated from German into French, Italian and English. A translation memory system, containing the translations of the avalanche bulletins of the last 15 years, helped to reduce the translation time. A comparison of this text corpus with the Canadian TAUM-Météo translation model [8] showed that the sentences collected over all those years cannot be expected to be comprehensive enough to directly extract a catalogue of phrases, or to be used for statistical machine translation (let alone for a system that does not require proofreading or manual corrections). For these reasons, a custom-made and fully automated translation system was built, which implements an approach based on a catalogue of standard phrases and has been in productive use since November 2012.

This kind of catalogue-based translation system has been used before, e.g. for severe weather warnings [15], but to our knowledge only for simpler domains and less complex sentence types. Our approach to create the catalogue was already described in [17], from where we summarize some content at the beginning of section 3 to give the relevant background and to show the peculiarities of our development. The end of section 3 and sections 4 and 5, form the main contribution of this paper, presenting for the first time systematic analyses of the system. These evaluations cover both, the possibilities given to authors with regard to content as well as the quality of the automatic translations, as compared to the old, manually written and translated danger descriptions.

## 2      Background

The languages generated by our catalogue system can be considered Controlled Natural Languages (CNL) [7]. The first CNLs with the goal to improve translation appeared around 1980, such as Multinational Customized English [14] and Perkins Approved Clear English [11]. Further languages were developed in the 1990s, including KANT Controlled English [9] and Caterpillar Technical English [5]. The goal was always to improve the translation by either making the work of translators easier by providing more uniform input texts or by producing automated translations of sufficient quality to be transformed into the final documents after manual correction and careful post-editing. Adherence to typical CNL rules has been shown to improve quality and productivity of computer-aided translation [1, 10]. For the Controlled Language for Crisis Management, it has been shown that texts are easier to translate and require less time for post-editing [16].

In contrast, the Grammatical Framework (GF) [12] is a general framework for high-quality rule-based machine translation. It is usable in narrow domains without the need for post-editing, such as the one presented in this paper. GF applies deep linguistic knowledge about morphology and syntax and has been used in prototypes such as AceWiki-GF [6] and a system enhanced by statistical machine translation to translate patents [3], but it does not yet have applications in productive use that match the complexity of texts of our problem domain of avalanche bulletins.

With PILLS [2], as a further comparable system, master documents containing medical information can be automatically transformed into specific documents for different user groups and translated into different languages. As the outcome of a one-year research project, PILLS was a prototype and was – to our knowledge – never applied operationally.

In terms of the PENS categorization scheme for CNLs [7], the languages presented here fall into the category P=2, E=2, N=5, S=4: They have relatively low precision and expressiveness (seen from a formal semantics point of view), but are maximally natural and comparatively simple.

## 3     Catalogue-based translation system

In this section, we give a summary of the methods which we described in our previous paper [17]. In general, a catalogue-based translation system is a collection of predefined phrases (or sentence templates) and therefore cannot be used to translate arbitrary sentences. The phrases in our system were created in the source language German, translated manually into the target languages French, Italian and English, and stored in a database. The editing tool for the creation of the phrases follows an approach similar to conceptual authoring [4, 13]: sentences are created by first selecting a general sentence pattern from a list and then gradually specifying and expanding the different sentence components. Once a phrase is chosen, it is immediately available in the target languages.

The individual sentences are not static but consist of a succession of up to ten segments. For each segment, the authors can select from a pull-down menu of predetermined options. These options can likewise consist of a series of sub-segments with selectable options, and, as part of the sub-segments, even sub-sub-segments are possible. Theoretically, the 110 predefined phrases could be used to generate several trillion different sentences. Not all possible sentences are meaningful, but all those that make sense must have correct translations in all languages. As no proofreading is possible in operational use, the translations in the catalogue must be guaranteed to be of high quality.

### 3.1     Creating the phrases in the source language

The sentences were created by an experienced avalanche forecaster whose native language is German and who has a good knowledge of all the target languages. Numerous avalanche bulletins from the past 15 years were consulted in order to cover as many situations as possible. No phrases were taken directly: their content was always generalized and the phrase structure was simplified wherever possible. The challenge was to find sentences that were universal enough to describe all the possible danger situations and simple enough to be translated. No explicit simplified grammar was used in any language. As a sentence can only be used when it works in the source language as well as in all target languages, the original German sentences had to obey the following rules:

- In each individual language, adjectives can only be used when they refer to subjects with the same gender and number in all the options.

- Articles depend on number and – in most of the languages used – gender and must therefore usually be included in the same option as the noun.
- Prepositions often change with the noun and must therefore also be included in the same option as the noun, e.g. 'in' Ticino (a region), but 'on the' Rigi (a mountain).
- As German has four grammatical cases, this sometimes necessitated splitting certain phrases into additional segments and sub-segments. E.g. "Fresh snow drifts require caution / are to be avoided" must be separated from "Fresh snow drifts represent the main danger", because in German the case of "fresh snow drifts" turns "Frische*n* Triebschnee beachten / umgehen" into "Frische*r* Triebschnee ist die Hauptgefahr".
- Demonstrative pronouns are only allowed to substitute one specific noun. Thus, e.g. the German "diese" is listed twice in the same pulldown, once for "the avalanches" (in Italian the feminine "queste ultime") and once for "the snow drifts" (in Italian the masculine "questi ultimi"). As there is no difference in the source language German, the substituted noun is indicated in the bulletin editor beside the pronoun, which allows the avalanche forecasters to choose correctly.

### 3.2 Translation of the catalogue

Translations take place on the segment level. Although German, French, Italian and English are all Indo-European languages, the differences in word order, gender, declension and so on make segmented translation difficult. Thus, specific editing and visualization software had to be developed by a translation agency to prepare the phrase translations. The translations themselves were performed manually by professional translators familiar with the topic and applying our text corpus. In addition to the omnipresent problem of inflection, ensuring the correct word order also proved difficult. Other problems included:

- clitics, apostrophes and elisions to avoid hiatus, especially in French and Italian;
- the Italian impure 's' ("*i grandi* accumuli" but "*gli spessi* accumuli");
- the split negation in French ("ne ... pas").

When translating the individual sentences and options, no logical functions, distinction of cases or post processing were used, except for a check to ensure the presence of a space between the different segments and a capital letter at the beginning of each sentence. In comparison with the source language, only two changes were allowed in the target languages (Fig. 2): (1) the segment order could vary between the languages (but is fixed for any given language and thus independent from the chosen options) and (2) each segment could be split in two (into ...a, ...b, Fig. 2). The latter facility was widely used, mainly to construct idiomatic word orders. This splitting is only used in the target languages and limits the use of our system to translations from German into the other languages but not backwards. Technically, the system could be used from any language to any other in the language matrix, but when producing the input it would be difficult for forecasters to find the correct sentences in a source language with segment splits.

Apostrophes, elisions, clitics and the impure 's' were handled by using pulldown splits or by taking all together into the same option. The latter required sometimes splits across the constituent units. As splits are invisible, this did not detract the output.

| Segment 1 | Segment 2 | Segment 3 | | Segment 4 | Segment 5 |
|---|---|---|---|---|---|
| die Lawinen | können | | | | gross werden. |
| nasse Lawinen | | auch | | oft | weit vorstossen. |
| diese | | {on_steep} Sonnenhängen | | weiterhin | bis in die aperen Täler vorstossen. |
| | | in diesen Gebieten | | | bis in tiefe Lagen vorstossen. |

| Segment 3a | Segment 1 | Segment 2 | Segment 3b | Segment 4 | Segment 5 |
|---|---|---|---|---|---|
| | the avalanches | can | | | reach large size. |
| | wet avalanches | | also | in many cases | reach a long way. |
| {on_steep} sunny slopes | they | | | as before | reach the bare valleys. |
| in these regions | | | | | reach low altitudes. |

**Fig. 2.** Schema of a phrase in the source language German (above). {on_steep} mark a sub-segment with several further options. In this example, [blank] is one of the options in the third and fourth segment. In English, the order of the segments is different and segment 3 is split.

### 3.3 Operational use

Since going operational, nearly 2000 danger descriptions have been produced per language. As before, the danger descriptions in German were proofread and discussed by at least two avalanche forecasters. Once the content of the German text was found to be correct, the translated texts were published without any further proofreading or corrections.

In a systematic survey we performed, all forecasters rated their satisfaction with the catalogue as "excellent". Six out of seven forecasters declared that at least "almost always" the differences between what they wanted to write and what they could write with the catalogues fell within the range of uncertainty regarding the current danger situation. "Greater limitations" never occurred. In the case of missing sentences, the system allows to add arbitrary text strings in all four languages and to use them immediately. However, no such 'joker phrases' were actually used during the first two winters of operational service.

## 4 Quality of the texts

### 4.1 Method

To assess the language quality, we compared in a blind study texts from old, manually written and translated descriptions with the new descriptions from the catalogue.

To get a comparable set, we chose one danger description from the evening edition of the avalanche bulletin from every second calendar day, starting at the beginning of December and finishing at the end of March. The descriptions from the catalogue were taken from the 2012/13 winter season, the freely written descriptions from the issues from winter 2011/12 back to 2007/08. To avoid evaluating texts that were too short, we only used danger descriptions with more than 100 characters in German. On days with more than one danger description, we randomly chose one of them.

**Table 1.** Questions concerning the language quality (correctness, comprehensibility, readability and clarity)

| 1. Is the text correct? <br> ("minor error" = typing mistake, incorrect punctuation or use of upper/lower case letters...) | | | | |
|---|---|---|---|---|
| Absolutely correct | 1 minor error | several minor / 1 major error | several major errors | Completely wrong |
| 2. Is the language easy to understand? (Assuming familiarity with the key technical terms) | | | | |
| Very easy to understand | Easy to understand | Understandable | Difficult to understand | Incomprehensible |
| 3. Is the text well formulated and pleasant to read? | | | | |
| Very well crafted | Easy to read | Clear | Difficult to read | Barely or not at all readable |
| 4. Is the situation described clearly? | | | | |
| Clearly and precisely | Reasonably clearly | Understandably | Unclearly, meaningless | Incomprehensibly, contradictory |

As 120 descriptions per language are too much for a survey, we divided them into 6 different sets, containing 10 descriptions from the old and new bulletin each. We divided the descriptions into the different sets in such a way that different avalanche situations were distributed as uniformly as possible. The order of the descriptions was chosen randomly for each dataset, but identical across all languages.

For every description, we asked four questions about the language quality (Tab. 1) and, additionally, in what manner the participant assumed the text was produced.

The survey was posted on www.slf.ch, on the website of the Swiss avalanche warning service, from 18 February to 5 March 2014. Each participant randomly received one out of the six data sets, in the same language as the website was visited. After a quality check, we had usable data from 204 participants.

93% of the participants were native speakers, 81% were men. The mean age was 43 years. Reflecting the languages of the visitors of our website, we received the most answers for German (76) and the least for English (18). With a median between "medium" and "high", English participants rated their experience in evaluating avalanche dangers slightly higher than the other participants with "medium". Other particularities when comparing the participants using the different languages were not found.

**Table 2.** Participants in the survey, divided into languages and allotted datasets

| Language, n per set (1/../6) | German, 76 <br> 14/11/13/10/16/12 | English, 18 <br> 3/2/3/5/4/1 | French, 55 <br> 10/12/9/6/6/12 | Italian, 55 <br> 9/9/7/12/8/10 |
|---|---|---|---|---|

### 4.2 Analysis

The age of the participants shows normal distribution permitting the use of the t-test. To analyse the detection rate of the origin of a description within a language, the data were cross-tabulated and the chi-square statistic was calculated. All differences between categorical variables were tested with the Mann-Whitney U-test for statistical significance (using $p = 0.05$).

When comparing the language quality of old and new descriptions, we could only find differences in isolated cases by using common parameters for ordinal data as median or mode. As we did not wish to jump to the conclusion that there was no difference at all, we assumed the predefined responses to be equal in distance and allocated numerical values to the different categories, starting with 5 for the best rating and 1 for the worst. We only used these numerical values to calculate mean values in order to show differences between different languages and between the old and the new descriptions.

Not all of the 6 datasets of a particular language had the same number of usable answers (Tab. 2). We therefore checked our data in every language for anomalies in distribution between the different datasets. As we did not find any, we pooled all the answers together.

To test significances between different languages, as well as to analyze the overall rating over all the languages, we used a balanced dataset. This contained all the English answers and in each of the other languages randomly chosen ratings of 180 descriptions from the old and the new bulletin each.

### 4.3 Results

The evaluators detected the origin of a given text in 59% of the German descriptions (Tab. 3). In the target languages, the rate of correct recognition was lower, with 55% in French and 52% in Italian and English. The recognition rate was significantly better than random only for German and French.

Table 4 shows answers to questions regarding the real origin of the danger descriptions. Differences between old and new descriptions are small and vary from language to language. Thus, with our balanced dataset we only get a significant decrease taking all languages and all questions together ($p = 0.02$), but not for individual questions.

**Table 3.** Correct ratings of the origin of the descriptions. Significant values are highlighted.

|  | German | English | French | Italian |
|---|---|---|---|---|
| n (equally balanced old/new) | 1520 | 360 | 1100 | 1100 |
| detection rate | **0.59** | 0.52 | **0.55** | 0.52 |
| p - value | p < 0.001 | p = 0.40 | p < 0.001 | p = 0.13 |

**Table 4.** Rating for the new descriptions from the catalogue of phrases and difference between new and old descriptions. Better ratings for the new descriptions are marked green, lower ratings red. Significant differences are highlighted. *are calculated from the balanced dataset.

|  |  | correct | comprehensible | readable | clear | all |
|---|---|---|---|---|---|---|
| German (n=1520) | mean | 4.75 | 4.30 | 3.93 | 4.29 | 4.32 |
|  | difference (new-old) | 0.03 (p=0.22) | **0.13 (p=0.003)** | 0.05 (p=0.25) | **0.16 (p=0.001)** | **0.09 (p<0.001)** |
| English (n=360) | mean | 3.89 | 3.74 | 3.51 | 3.73 | 3.72 |
|  | difference (new-old) | -0.01 (p=0.61) | 0.01 (p=0.90) | 0.03 (p=0.95) | -0.05 (p=0.45) | -0.003 (p=0.54) |
| French (n=1100) | mean | 4.57 | 4.30 | 4.07 | 4.34 | 4.32 |
|  | difference (new-old) | **-0.12 (p=0.001)** | -0.04 (p=0.42) | **-0.11 (p=0.01)** | 0.01 (p=0.47) | **-0.07 (p=0.001)** |
| Italian (n=1100) | Mean | 4.35 | 4.21 | 3.99 | 4.28 | 4.21 |
|  | difference (new-old) | **-0.16 (p=0.001)** | -0.09 (p=0.08) | -0.08 (p=0.12) | **-0.12 (p=0.01)** | **-0.11 (p<0.001)** |
| all languages | mean | 4.39 | 4.14 | 3.87 | 4.16 | 4.14 |
|  | difference (new-old) | -0.06 (p=0.08)* | 0.004 (p=0.21)* | -0.03 (p=0.10)* | 0.001 (p=0.08)* | **-0.02 (p=0.02)*** |

## 5 Discussion

According to the avalanche forecasters, the catalogue of phrases always allowed an adequate description of the danger situation. The translations in the catalogue were checked extensively by the developer, an experienced avalanche forecaster with knowledge in all four languages. The catalogue proved to be even more exact with regard to content, as the manual translation method used for old avalanche bulletins lacked the necessary time to correct smaller inconsistencies.

The detection rate was statistically significant above the random value, but the number of correctly recognized descriptions was small with 55% on average for all languages. This corresponds to, for example, correctly recognizing 2 out of 20 descriptions and then tossing a coin for the remaining 18 descriptions.

French and German speaking participants rated the language quality best with an overall value of 4.32. Italian was nearly as good with a value of 4.21. English ratings were significantly lower and this in both, the old and the new descriptions with a mean of 3.75 and 3.72 respectively. Perhaps this is due to the fact that our translators are British and by using the glossary of the European Avalanche Warning Services (www.avalanches.org) which differs substantially from terminology used in North America, where at least some of the participants of the survey live.

In addition, the large variance between different participants in assessing the same dataset shows that the absolute value of the rating is not only a question of the sentences, but also possibly affected by varying interpretations of the given texts or some other habit of the individual participant. To understand this anomaly, further research would be needed. In our survey, we are much more interested in the changes in language quality due to the introduction of the catalogue of phrases than in the absolute value of quality. Our purpose is hardly affected by these anomalies, thanks to a symmetrical dataset with always contains the same number of descriptions from each origin.

Compared to the differences between the languages, the differences between old and new descriptions are small. This is surprising because the introduction of the catalogue of phrases was a fundamental change and the catalogue itself was mostly translated by different translators.

Of all properties, the correctness reaches the best rating (Tab. 4), whereas the comprehensibility and the clarity of the formulations lie ex aequo in the middle of the investigated parameters about the language quality. The catalogue of phrases leads to a standardized language. As Swiss avalanche forecasters believe that this kind of simple and unambiguous language is well suited to communicate warnings, they wrote the "old" danger descriptions in a similar way as well. In this context, it is not surprising that of all the quality criteria, the pleasure to read was assessed lowest.

In German the descriptions generated with the catalogue of phrases were rated even better, for all four criteria (correctness, comprehensibility, readability and clarity). Note that German is the source language of both, the manually written texts and the catalogue of phrases. In Italian and French, descriptions from the catalogue of phrases were rated lower, for all four criteria in Italian and for three criteria in French. However, the differences are small and in many cases not significant (Tab. 4). In English, no noteworthy change was found.

Statistical significance is a question related to the number of trials, and with the more than 5,500 assessments used in our balanced dataset we can test for even slight differences. Consequently, the decrease of the language-weighted mean values over all questions is statistically significant, even when numerical values for our ordinal data shows, that the decrease was marginal with a value falling from 4.16 to 4.14.

Given the only marginal change in ratings from old to new, and the poor recognition rate, we conclude that in the majority of cases, users did not notice a decrease in the language quality with the introduction of the catalogue. Thus, the language quality from the catalogue of phrases can be judged as virtually equivalent to the text written from scratch and translated by topic-familiar professionals.

## 6    Conclusions

The catalogue-based system proved to be well-suited to generate the Swiss avalanche bulletin. After two years of operational use, all forecasters declared that within the limited time available to produce forecasts, it was possible to describe the different avalanche situations with precision and efficiency.

The system also proved to be well-suited for fully automatic and instantaneous translation of danger descriptions from German into the target languages French, Italian and English. The translations do not need to be proofread or corrected, and they turned out to be even better with respect to their content than the manual translations of the old avalanche bulletin.

The quality of the language was assessed in a blind study, comparing old, manually translated danger descriptions with new, catalogue-based danger descriptions. Recognizing the difference proved to be difficult; the mean detection rate was only 55%. Based on four criteria the quality of danger description was rated good with some differences between languages. With the introduction of the catalogue of phrases, there were only marginal changes in the different quality ratings. Depending on the language, they show a small improvement or a slight decrease in quality. Thus the bulletins produced by the catalogue of phrases were virtually equivalent in language quality to those produced using the old method of ad hoc translation.

As using a phrase catalogue requires experience, frequent operational use is necessary. It is crucial that users find the phrases matching the given danger situation quickly enough, which has shown to be the case for our system. The implemented search engine was essential. Our experience has shown that the number of phrases should be kept to a minimum by reusing individual phrases in multiple contexts, and that the presented approach is particularly well-suited if the problem domain can be described by a small sublanguage, as is the case for the highly specific topic of avalanche forecasting.

With respect to financial aspects, the cost-benefit ratio of our system turned out to be excellent. The savings from not needing manual translations are expected to exceed the initial development costs within a few years. Applying the database to other multi-lingual countries or extending it to topics such as weather forecasting is conceivable. An adaption to very different languages seems difficult due to differences in grammar and language usage.

The construction of the catalogue and the translations had both been done in an empirical way. We gladly place them at the disposal for further investigations.

## 7   Acknowledgments

We thank Martin Bächtold from www.ttn.ch and his translators for the courage to translate the catalogue of phrases and for maintaining a high quality standard. Furthermore, we thank Eva Knop, Nico Grubert, Frank Techel and Curtis Gautschi for valuable input and the participants of the survey.